\documentclass[a4paper,12pt,fleqn]{cas-sc}

\usepackage[authoryear]{natbib}

\usepackage{hyperref}

\usepackage{xurl} 
\usepackage{array}
\usepackage{makecell}
\usepackage{caption}

\usepackage[left]{lineno} 

\def\tsc#1{\csdef{#1}{\textsc{\lowercase{#1}}\xspace}}
\tsc{WGM}
\tsc{QE}
\tsc{EP}
\tsc{PMS}
\tsc{BEC}
\tsc{DE}

\renewenvironment{abstract}{}{}

\renewcommand{\abstractname}{}
\AtBeginDocument{}

\makeatletter
\def\@abstract{}          
\def\@elsarticle@abstract{} 
\makeatother

\begin{document}
\let\WriteBookmarks\relax
\def\floatpagepagefraction{1}
\def\textpagefraction{.001}
\raggedbottom



\title [mode = title]{Machine Learning Detection of Road Surface Conditions: A Generalizable Model using Traffic Cameras and Weather Data}            

\author[1,2]{Carly Sutter}[
                        orcid=0009-0008-1414-4386]

\cormark[1]


\ead{csutter@albany.edu}

\credit{Conceptualization, Data curation, Formal analysis, Investigation, Methodology, Software, Visualization, Writing – original draft}

\affiliation[1]{organization={Atmospheric Sciences Research Center, University at Albany},
    city={Albany},
    state={NY},
    postcode={12222}, 
    country={USA}}

\author[1]{Kara J. Sulia}[orcid = 0000-0002-4054-4956]
\cormark[1]
\ead{ksulia@albany.edu}
\credit{Conceptualization, Funding acquisition, Methodology, Resources, Supervision, Writing – review \& editing}

\author[1,3]{Nick P. Bassill}[orcid=0000-0002-9105-0761]
\credit{Conceptualization, Funding acquisition, Methodology, Resources, Writing – review \& editing}

\author[4]{Christopher D. Wirz}[orcid=0000-0002-8990-5505]
\credit{Conceptualization, Data curation, Investigation, Methodology, Writing – review \& editing}

\author[1]{Christopher D. Thorncroft}[orcid = 0000-0001-7872-6459]
\credit{Conceptualization, Funding acquisition, Resources, Writing – review \& editing}

\author[5]{Jay C. Rothenberger}[orcid = 0009-0007-2530-4667]
\credit{Conceptualization, Investigation, Methodology, Writing – review \& editing}

\author[1]{Vanessa Przybylo}[orcid=0000-0003-4380-3543]
\credit{Conceptualization, Data curation, Investigation, Methodology, Writing – review \& editing}

\author[4]{Mariana G. Cains}[orcid=0000-0002-6729-6729]
\credit{Data curation, Investigation, Methodology, Writing – review \& editing}

\author[6]{Jacob Radford}[orcid=0000-0001-6824-8967]
\credit{Data curation, Investigation, Writing – review \& editing}

\author[1,2]{David Aaron Evans}[orcid=0009-0003-1537-4996]
\credit{Data curation, Investigation, Writing – review \& editing}

\affiliation[2]{organization={Department of Atmospheric and Environmental Sciences, University at Albany},
    city={Albany},
    state={NY},
    postcode={12222}, 
    country={USA}}

\affiliation[3]{organization={State Weather Risk Communication Center},
    city={Albany},
    state={NY},
    postcode={12222}, 
    country={USA}}


\affiliation[4]{organization={National Science Foundation National Center for Atmospheric Research},
    city={Boulder},
    state={CO},
    postcode={80305}, 
    country={USA}}
    
\affiliation[5]{organization={School of Computer Science, The University of Oklahoma},
    city={Norman},
    state={OK},
    postcode={73019}, 
    country={USA}}
    
\affiliation[6]{organization={Cooperative Institute for Research in the Atmosphere},
    city={Fort Collins},
    state={CO},
    postcode={80521}, 
    country={US}}
    
\cortext[cor1]{Corresponding author}

\markboth{\__short_authors:}{}



\noindent
\small \textit{© 2026. This manuscript version is made available under the CC-BY-NC-ND 4.0 license \url{https://creativecommons.org/licenses/by-nc-nd/4.0/}}

\vspace{1em}
\noindent
\small \textit{This is the author's version of a work that has been accepted for publication in the International Journal of Transportation Science and Technology. Changes were made to this version based on peer review. A URL/DOI to the final version will be provided when available.}
\normalsize

\begin{center}
{\LARGE Machine Learning Detection of Road Surface Conditions: A Generalizable Model using Traffic Cameras and Weather Data \par}
\vspace{1em}

Carly Sutter\textsuperscript{1,2,*} ORCiD: 0009-0008-1414-4386 \\
Kara J. Sulia\textsuperscript{1,*} ORCiD: 0000-0002-4054-4956 \\
Nick P. Bassill\textsuperscript{1,3} ORCiD: 0000-0002-9105-0761 \\
Christopher D. Wirz\textsuperscript{4} ORCiD: 0000-0002-8990-5505 \\
Christopher D. Thorncroft\textsuperscript{1} ORCiD: 0000-0001-7872-6459 \\
Jay C. Rothenberger\textsuperscript{5} ORCiD: 0009-0007-2530-4667 \\
Vanessa Przybylo\textsuperscript{1} ORCiD: 0000-0003-4380-3543 \\
Mariana G. Cains\textsuperscript{4} ORCiD: 0000-0002-6729-6729 \\
Jacob Radford\textsuperscript{6} ORCiD: 0000-0001-6824-8967 \\
David Aaron Evans\textsuperscript{1,2} ORCiD: 0009-0003-1537-4996 \\

\vspace{0.5em}

\textsuperscript{1}Atmospheric Sciences Research Center, University at Albany, Albany, NY 12222, USA \\
\textsuperscript{2}Department of Atmospheric and Environmental Sciences, University at Albany, Albany, NY 12222, USA \\
\textsuperscript{3}State Weather Risk Communication Center, Albany, NY 12222, USA \\
\textsuperscript{4}National Science Foundation National Center for Atmospheric Research, Boulder, CO 80305, USA \\
\textsuperscript{5}School of Computer Science, The University of Oklahoma, Norman, OK 73019, USA \\
\textsuperscript{6}Cooperative Institute for Research in the Atmosphere, Fort Collins, CO 80521, USA \\

\vspace{0.5em}

* Corresponding authors: Carly Sutter (csutter@albany.edu), Kara J. Sulia (ksulia@albany.edu)
\end{center}

\clearpage


\section*{Abstract}

Transportation agencies make critical operational decisions during hazardous weather events, including assessment of road conditions and resource allocation. In this study, machine learning models are developed to provide additional support for the New York State Department of Transportation (NYSDOT) by automatically classifying current road conditions across the state. Convolutional neural networks and random forests are trained on NYSDOT roadside camera images and weather data to predict road surface conditions. This task draws critically on a robust hand-labeled dataset of \(\sim22{,}000\) camera images containing six road surface conditions: severe snow, snow, wet, dry, poor visibility, or obstructed. Model generalizability is prioritized to meet the operational needs of the NYSDOT decision makers, including integration of operational datasets and use of representative and realistic images. The weather-related road surface condition model in this study achieves an accuracy of 81.5\% on completely \textit{unseen} cameras. With operational deployment, this model has the potential to improve spatial and temporal awareness of road surface conditions, which can strengthen decision-making for operations, roadway maintenance, and traveler safety, particularly during winter weather events.

\section*{Keywords}
\begin{itemize}
    \item Machine learning
    \item Traffic camera imagery
    \item Winter road maintenance
    \item Co-design 
    \item New York State Department of Transportation
\end{itemize}

\section*{Highlights}
\begin{itemize}
\item Developed a model to classify road surface conditions using image and weather data
\item Achieved accuracy of 81.5\% on completely unseen cameras for weather-related classes
\item Integrated co-design with end-users and interdisciplinary collaboration
\item Designed methods that prioritize model generalizability for operational applicability
\end{itemize}

\clearpage  


\section{Introduction}





Intelligent systems and use of data-driven approaches have been critical for many parts of the transportation sector, including traffic safety \citep{diverging}, traffic flow prediction \citep{trafficflow}, construction of new roadways, preventative maintenance of bridges, traffic management, urban mobility and sustainability, and public transportation \citep{OlivieraAvenues,OliveiraBridges,OlivieraUrban,OliveiraPublicTransportation,OlivieraTraffic}. Technology has played an essential role in developing unique solutions for transportation, and this work focuses on detecting road surface conditions that could support operational decision-making and maintenance, especially during winter months.

Inclement weather affects traveler safety, with dangerous weather conditions leading to more than 1 million accidents in the U.S. each year \citep{ncar-road-conditions}. Weather and road surface conditions have been shown to impact driver habits \citep{Liu}, affect traffic and accidents \citep{Park}, and therefore, affect traveler safety. The Federal Highway Administration \citep{fha} reported that over a 10-year period (2007 to 2016), 21\% percent of vehicle crashes were weather-related. With high snow rates in New York State (NYS), the impacts were even greater. From 2010 to 2019, about 35\% to 50\% of crashes along the NYS Thruway in the month of January were associated with inclement weather, especially snowfall \citep{CallFlynt}. Although fog was less frequent and accounted for only 3\% of weather-related crashes in the US, it accounted for 9\% of weather-related fatalities \citep{fha}. These statistics indicate the impact of road surface conditions on driver safety, motivating the work completed in this study.

\begin{figure}
\includegraphics[width=5.5in]{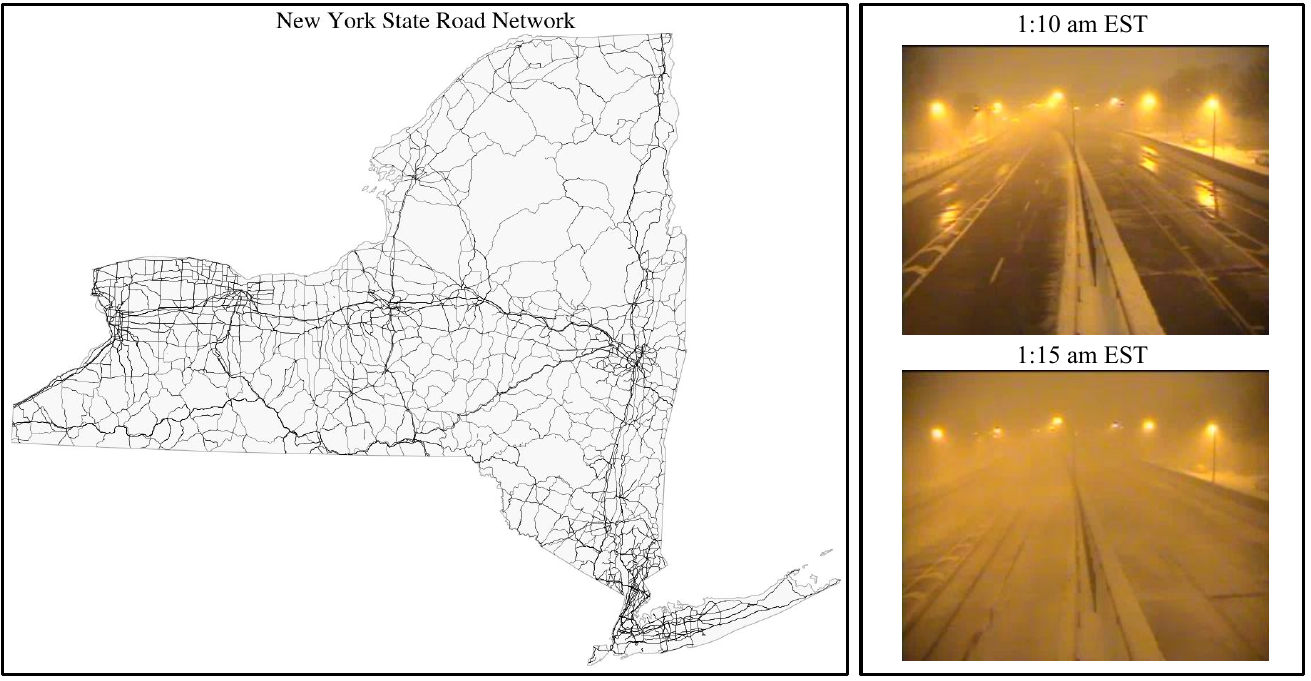}
\caption{\textbf{Roadways and motivation}. (Left) Topological road network graph of New York State. (Right) Example of one camera in Buffalo, NY on November 19, 2022, with two images taken five minutes apart, demonstrating how quickly snow can accumulate on roads.}\label{fig:roads}
\end{figure}

Awareness of road surface conditions is essential for transportation agencies to know where resources (e.g., snowplows, treatments) are needed. The New York State Department of Transportation (NYSDOT) has a network of approximately 2400 live traffic cameras across NYS, which are situated at varying heights and pointing toward the roadways. These cameras (referred to as ``sites” in this study) are publicly available at 511ny.org \citep{ny511} and winter road conditions are populated by the NYSDOT. Currently, the NYSDOT relies on manually observing individual cameras and driving on roads to determine the road surface conditions, both of which are time-intensive given the extensive road network across NYS (Fig. \ref{fig:roads}, left). However, machine learning (ML) models, such as convolutional neural networks (CNNs) and random forests (RFs), can be used to automate the task of road surface condition classification. Compared to manual methods, the benefit of using ML is that predictions are made frequently (e.g., every five minutes) and at the individual camera level, which provides a granular spatial breakdown. When direct observation is not available, the integration of ML could enable more efficient and effective decision making because the model could automatically identify areas where roads are snow-covered and quickly accumulating (Fig. \ref{fig:roads}, right). The purpose of this work is to use ML to make skillful predictions of road surface conditions on \textit{unseen} camera sites (those not used in model training), prioritizing the operational needs of the NYSDOT. 

\subsection{Literature review} \label{lit}

Camera images are used in transportation research across a variety of applications, using a range of datasets and methods. Many earlier studies \citep{Yamada2005,Kawai2014,ZhangChunyu2012,Sun2013,hunjunyang2014,Lee2016,kawarabuki2014} employ methods of extracting features from images such as texture, edges, or hue, saturation, and value, followed by  classification algorithms including  K-nearest neighbors, Naïve Bayes, and Neural Networks to predict various road surface conditions, achieving accuracies ranging 87\%-96\%.  A number of studies, including \cite{Yan2009}, \cite{omer2010}, \cite{qian2016}, \cite{pan2018}, \cite{wu2022}, use in-vehicle dash cameras from cars, buses, or snow plows to predict weather and road conditions. Various algorithms are used, such as Support Vector Machines and CNNs, achieving accuracies ranging 79\%-93\%. Other related studies incorporate information beyond road surface conditions, for example, \cite{Huang} use algorithms such as MonoDepth and XGBoost to predict different types of fog from roadway video data, \cite{ZHANG2022298} use CNNs to detect pavement distress like cracks and potholes, \cite{ports} develop a novel algorithm for autonomous navigation in ports, and \cite{sirirattanapol} use multi-class CNN to predict multiple attributes including traffic levels,  lighting, raining, and wet/dry road conditions. Road Weather Information System (RWIS) sensor data and camera images are used in \cite{jonsson} and \cite{carrillo2020design} to predict road surface conditions such as wet, dry, icy, and various levels of snow coverage, both works achieving accuracies above 90\%. Differences in underlying datasets, classes, and applications hinder direct comparison between these past studies, however, they demonstrate that automatic road surface condition detection has been a consistent and important problem over time.

Past studies that are most similar to the work presented in this paper include  \cite{walker2022}, \cite{carrillo2020integration}, and \cite{KHAN2022}. Specifically, \cite{walker2022} used a dataset of 1000 NYSDOT images to predict ``snow” or ``no snow” conditions using Google's Teachable Machine \citep{teachablemachine}, achieving test accuracy of 75\% for the ``snow” class and 95\% for the ``no snow” class. They also correlated image conditions with crash reports, finding that the chance of a crash increased by 61\% when ``snow” was predicted by the classification model. \cite{carrillo2020integration} discussed drawbacks to using RWIS data, including their cost, which limits the number of RWIS sites, and suggests incorporation of the Ministry of Transportation of Ontario cameras as an important research consideration. \cite{KHAN2022} used roadside images from Wyoming DOT CCTV cameras to predict road surface conditions. \cite{KHAN2022} also discussed the benefits of using already-available DOT images, stating that they are more prevalent and readily available than RWIS sites given that webcams are more cost effective. They hand-labeled their image dataset with a weather condition (clear, heavy snow, and light snow) and a road surface condition (dry, snowy, and wet/slushy), and then built a separate classification model for each, achieving at least 97\% validation accuracy on both. \cite{KHAN2022} also recognized the importance of testing their model on other roadways, so they tested their model on a small subset of 300 completely new cameras, achieving a high accuracy of 95.3\%. Their work does not include poor visibility or obstruction, and the importance of these classes are included in Section \ref{data}; more comparisons to \cite{KHAN2022} and the work presented in this study are discussed in Section \ref{scm}.

\subsection{Contributions}

This study focuses on methods for building a trustworthy model \citep{bostrom2023trust} with NYSDOT that can generalize across scenarios and locations. While many past studies addressed automated road-weather condition classification, they were often limited by few camera sites, few surface classes, limited dataset and labeling details, same-camera evaluation, or reliance on in-vehicle cameras or expensive sensors. Relative to past studies, this work offers five key contributions:

\begin{enumerate}\label{contributions}
    \item Close collaboration with the NYSDOT with the goal of creating a model that is trustworthy, a concept referred to as ``co-design” \citep{bostrom2023trust}. 
    \item Dataset curation integrates social science techniques and risk communication perspectives, ensuring transparent rules for human labeling that enhance understanding and can be applied to new camera systems.
    \item Emphasis on improving model performance on \textit{completely new} camera sites. Generalizability is a critical aspect to consider in order to eventually operationalize use of the model by the NYSDOT and other transportation agencies. 
    \item Creation of a robust dataset containing 45 unique cameras sites of \textit{varying quality} that are representative of cameras across the state, rather than including only high quality images that inflate performance. 
    \item Inclusion of varied classes that encompass a wide range of road surface conditions that are prevalent in images and critical for operational needs, developed in collaboration with NYSDOT. 
\end{enumerate}

\subsection{Overview}

\begin{figure}
 \centerline{\includegraphics[width=5.5in]{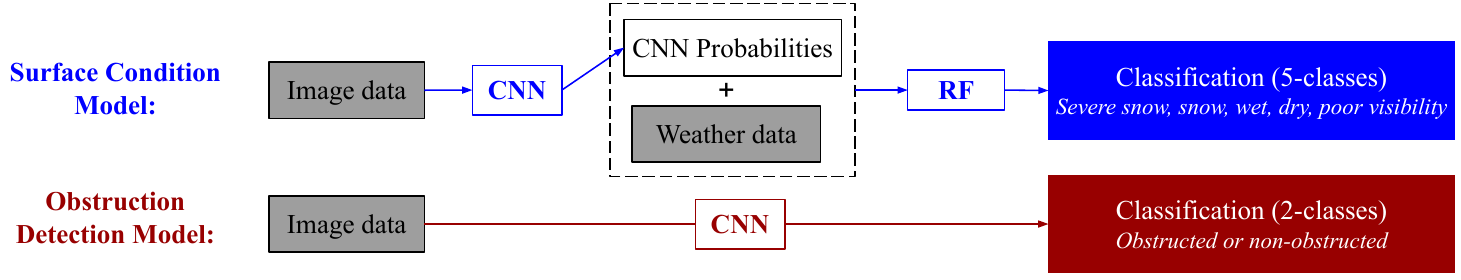}}
 \caption{\textbf{Operational model flow and example}. The Surface Condition Model is a two-stage process using a CNN and a downstream model (RF). The Obstruction Detection Model predicts whether an image is obstructed or non-obstructed. In a practical inference scenario, both predictions will be provided in conjunction for each new observation. }\label{fig:general_model_flow_alt}
\end{figure}

In this work, six road surface condition classes (severe snow, snow, wet, dry, poor visibility, and obstructed) are detected from camera images and weather data using supervised ML models. As shown in Figure \ref{fig:general_model_flow_alt}, the workflow consists of a Surface Condition Model, which predicts weather-related classes using a CNN followed by an RF, and an Obstruction Detection Model using a CNN. Section \ref{data} describes the hand-labeled image dataset and HRRR weather forecast data, and Section \ref{methods} outlines model training and selection. Section \ref{results} evaluates model performance, with particular attention to performance on unseen camera sites, referred to as ``generalizability”. Generalizability is critical for meeting operational needs that involve deploying the road surface classification models on thousands of cameras across NYS.

\section{Data} \label{data}
\subsection{Image and Weather Data}

A labeled dataset has been developed for this study for the development of ML models. This dataset consists of hand-labeled classifications of road surface conditions, which came from human labelers who manually viewed camera images and assigned a classification to each. Following a systematic approach known as Quantitative Content Analysis \citep{qcaorig}, this process included the development of a codebook containing documentation and rules for labeling each class \citep{Sutterzenodo}, where definitions were derived in consultation with the NYSDOT; for example, the presence of tire tracks distinguish between snow and severe snow. Critically, this labeling process also included a set of labeling trials where six people independently labeled the same set of images to ensure  reliability, and replicability of this hand-labeled dataset. Comprehensive details regarding the dataset for this study are published and available at \cite{Wirz2024}, \cite{Sutterzenodo}, and \cite{Sutterdatasheet}. A total of 21,653 images from 45 unique camera sites across NYS are labeled for this study. Table \ref{tab:examples_classes} shows examples of the six labeled classes, and detailed labeling definitions are in the codebook \citep{Sutterzenodo}.

A large focus of this work is to create a model that performs well on a wide range of conditions that are realistically encountered in the installed cameras, such as lower quality cameras (right-hand side of  Table \ref{tab:examples_classes}). Additionally, the obstructed cases seem less important because they are generally not weather related, but are included because they are observed frequently and can deem the observation as unreliable. While past studies may have included difficult examples like these, they have not been emphasized, and the focus of this work is to provide a model that can work on representative images from already-implemented cameras, which include these less-than-ideal cases. Building one model that can predict road surface conditions across a wide range of conditions on all cameras (including low quality cameras), creates a tool that is operationally realistic for the NYSDOT.

\begin{table}[h] 
\centering
\caption{The six road surface conditions are shown with examples and count distribution, totaling to 21,653 images. The right-most column shows examples of high- and low- quality images. The high-quality image is well lit even in the absence of car headlights, making the classification of road surface conditions more obvious. The low-quality image is poorly lit and is distanced far from the roadway, making the classification of these images more difficult; in these cases, the labeler relies on slight changes in road pavement darkness an observational weather data \citep{Sutterzenodo}.}
\label{tab:examples_classes}
\begin{tabular}{|p{0.22\linewidth}|p{0.22\linewidth}|p{0.22\linewidth}||p{0.22\linewidth}|}
\hline
\multicolumn{3}{|c||}{\textbf{Six classes}} & \multicolumn{1}{c|}{\textbf{Site Quality}} \\
\hline
Severe snow - 1,527 & Snow - 2,989 & Wet - 8,458 & High Quality - 58\%\\
\includegraphics[width=\linewidth]{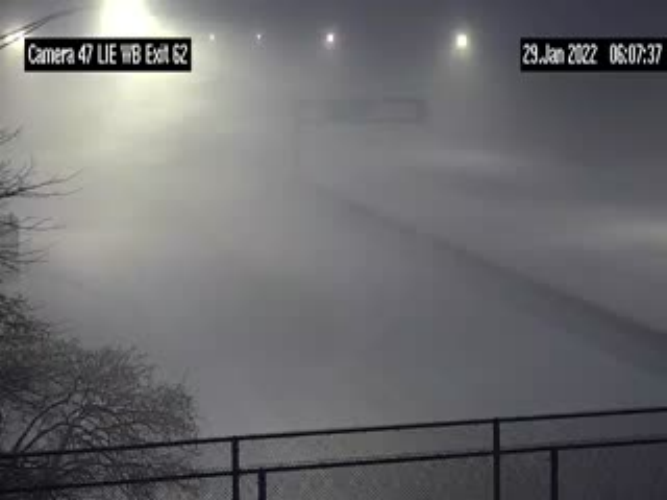} &
\includegraphics[width=\linewidth]{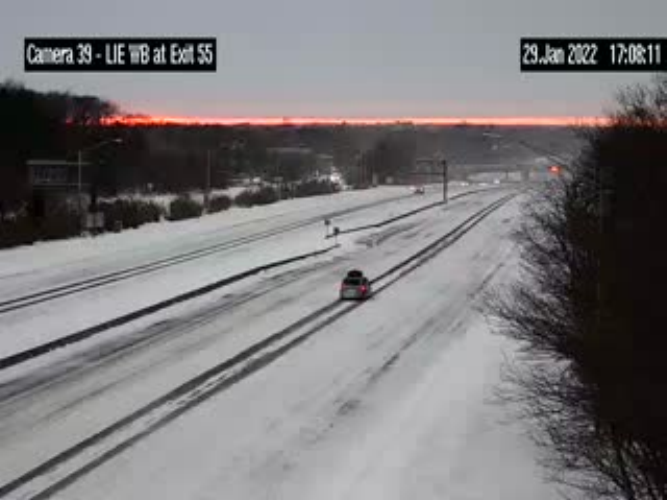} &
\includegraphics[width=\linewidth]{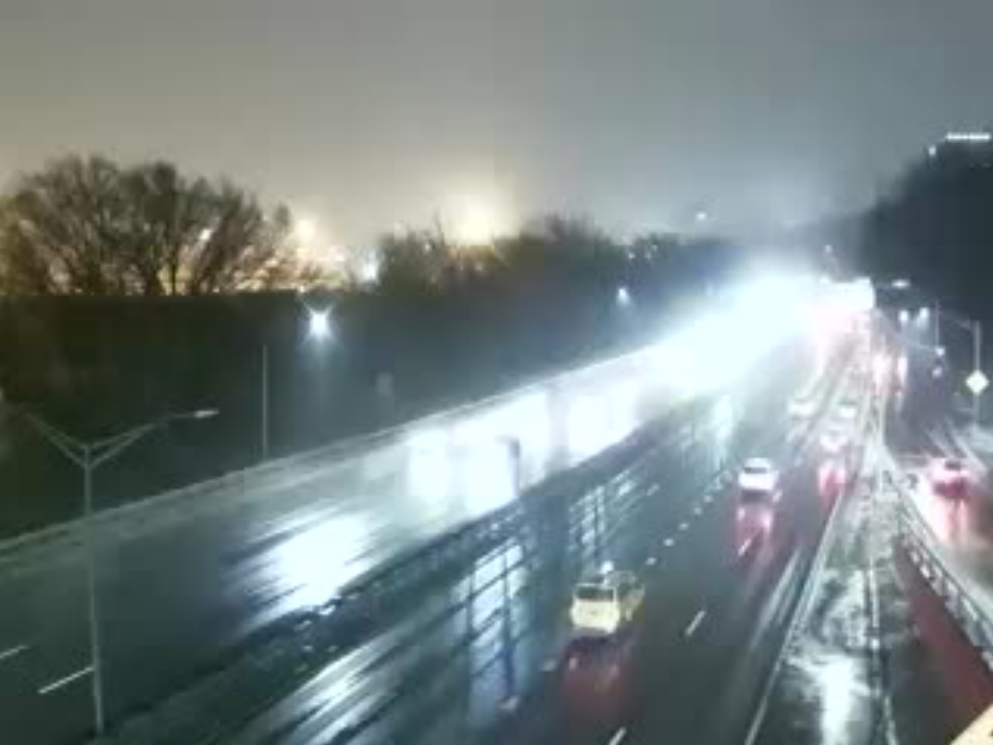} &
\includegraphics[width=\linewidth]{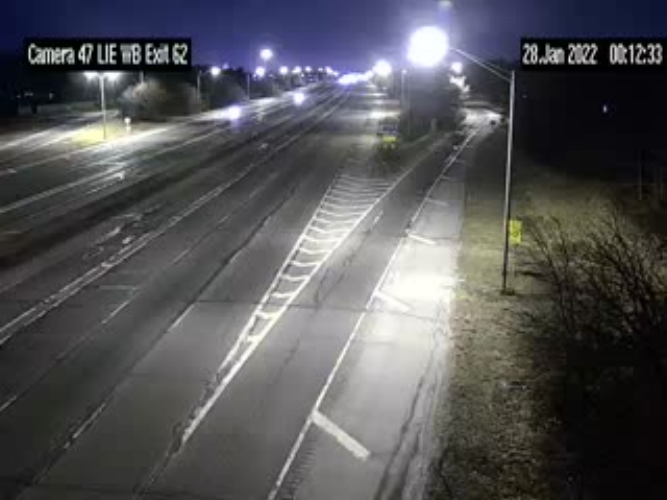} \\
\hline
Dry - 7,600 & Poor visibility - 915 & Obstructed - 164 & Low Quality - 42\%\\
\includegraphics[width=\linewidth]{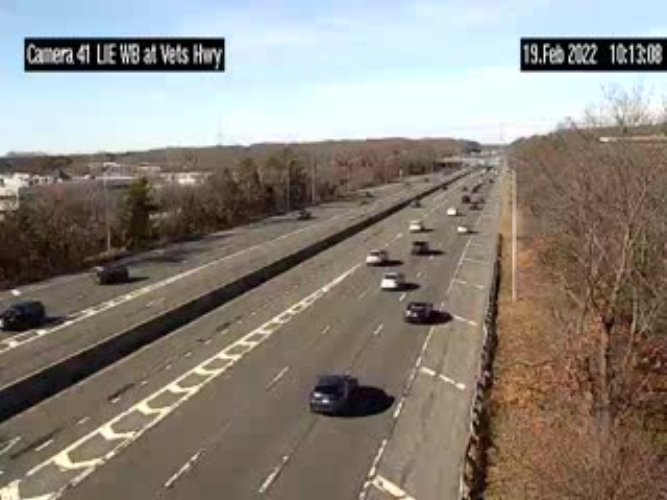} &
\includegraphics[width=\linewidth]{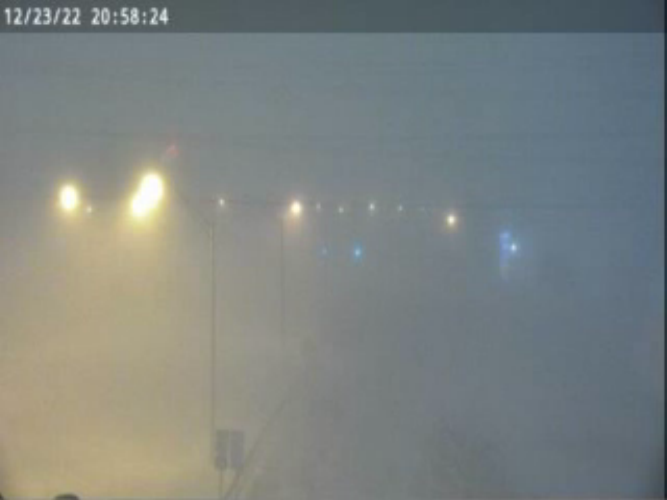} &
\raisebox{-.015\height}{\includegraphics[width=\linewidth]{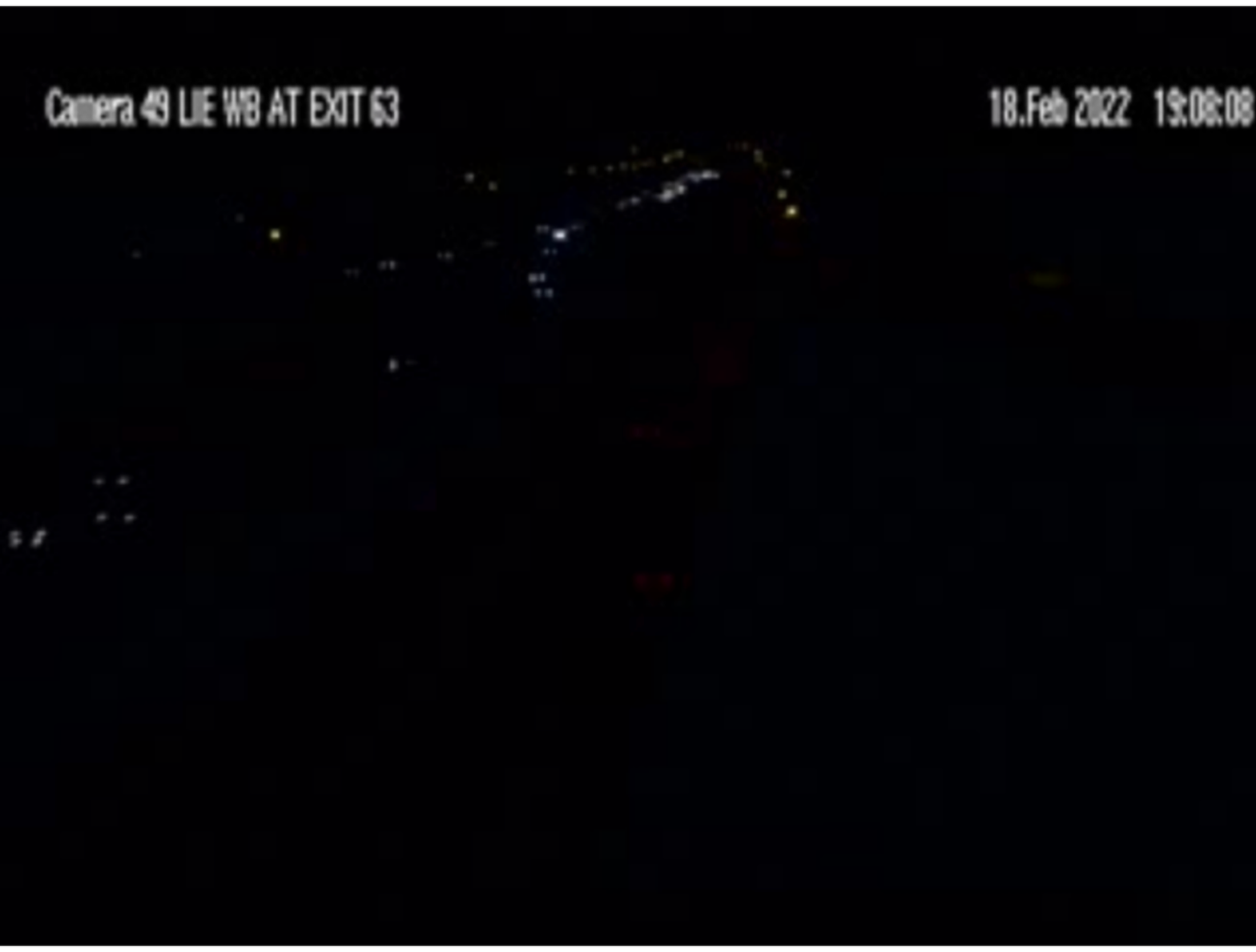}} &
\includegraphics[width=\linewidth]{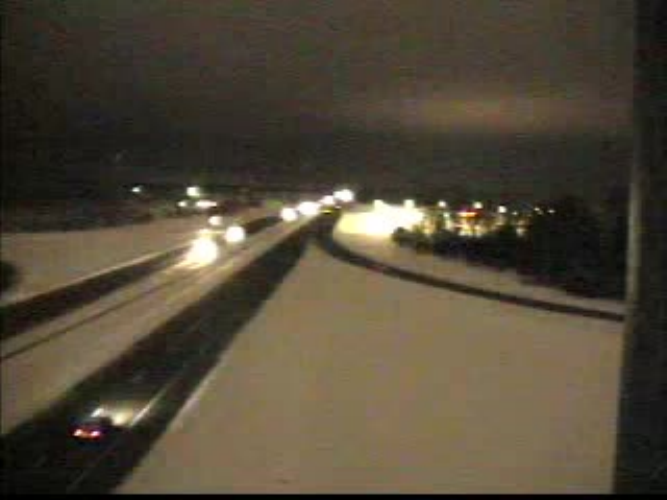} \\
\hline
\end{tabular}
\end{table}


The High-Resolution Rapid Refresh (HRRR) \citep{hrrr} is a weather forecast model specifically designed to meet the demands of applications that depend heavily on timely and accurate short-range meteorological data \citep{hrrr2013}. The labeled image data and forecast data are collated based on time and location resulting in corresponding weather data for each labeled image. Although forecasts contain errors compared to observational weather stations, such as the New York State Mesonet (NYSM) \citep{nysm}, forecast data is selected for this study for multiple reasons. First, the HRRR has a fine spatial grid with an average of 3km between HRRR grid points, whereas for the NYSM data, the average spacing is 27km. Second, with the broad goal to expand to other states and regions, the HRRR has large spatial coverage available nationwide, whereas relying on states to have expensive weather-observation stations could limit expansion and applicability. Lastly, the HRRR model uses frequent data assimilation to incorporate recent observations such as radar, producing highly reliable short-term forecasts \citep{gaudet}, and this study capitalizes on that reliability by utilizing a short lead time of 2 hours while also avoiding complications associated with initial spinup. Overall, weather data makes a positive contribution to the Surface Condition Model, resulting in an overall accuracy improvement of  +12.1\% (Sections \ref{scm} and \ref{gen}).

Six near-surface HRRR variables are used in this study including 2-meter temperature, 2-meter relative humidity, 10-meter average wind speed, 2-hr accumulated snow depth, 2-hr total precipitation, and total cloud cover. These six variables are selected based on their prominent relevance to road-surface weather and are similar to those used in \cite{carrillo2020design}. The HRRR data are not used to predict obstructions given that the obstructed class is generally not weather-related.

\subsection{Data splitting} \label{datasplitting}

\begin{figure}
 \centerline{\includegraphics[width=5.5in]{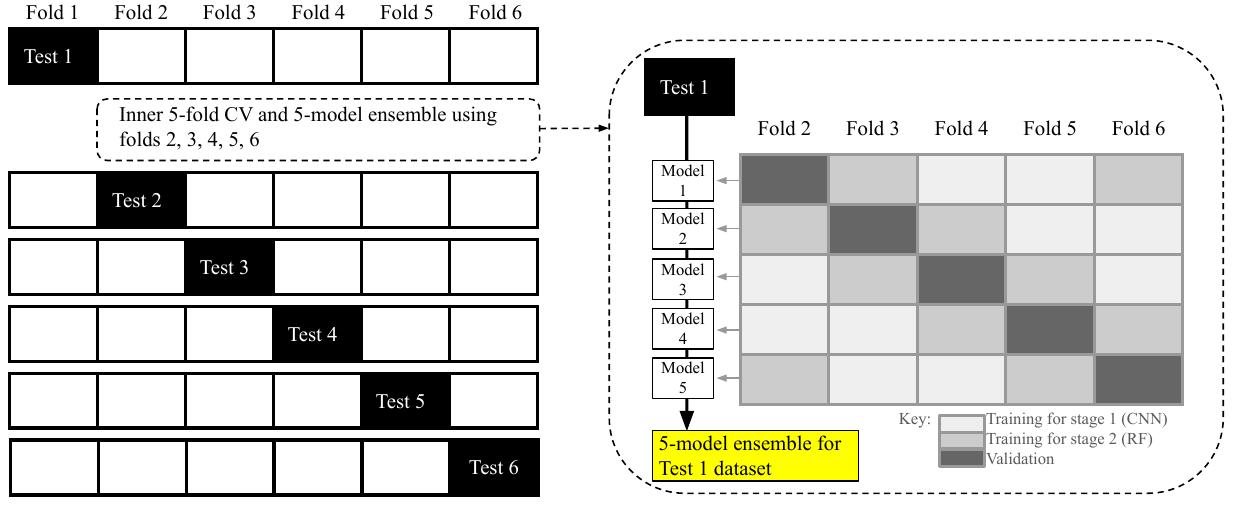}}
 \caption{\textbf{Nested CV for the Surface Condition Model}. For each test dataset (black cells shown on the left), inner 5-fold CV is used on the remaining five folds (right), resulting in five models. The five models are used to create an ensemble prediction for the test dataset. This process is demonstrated for Test 1 and is repeated for each test set.}\label{fig:nestcv}
\end{figure}

The data is split to ensure that model evaluation aligns with the goal of generalizability; this is critical given that labeling images from every camera site is not practical. Specifically, a ``site-specific” split method is used where all observations from a site are in the same fold, which ensures that training, validation, and testing contain unique sites. Folds are created to optimize the distribution of multiple attributes, although exactly equal composition is not feasible given the iterative nature of labeling \citep{Sutterdatasheet}. Specifically, each of the six folds are comprised of between 1) 22-29\% hazardous classes (severe snow, snow, or poor visibility), 2) 6-8 unique sites, 3) 38-55\% low quality sites 4) 70-90\% from the months of January or February, and 5) 52-56\% night instances. Winter months are prioritized for operational purposes, but the same model works in summer months, predicting wet, dry, and poor visibility classes. Data splitting for the Surface Condition Model is demonstrated visually in Fig. \ref{fig:nestcv}, showing how nested cross-validation (CV) is used to ensure that each stage of the modeling process is evaluated on data not used in model training. The labeled data is split into six folds of approximately equal size (16.67\%) that serve as the training, validation, or testing data at different points in the modeling process. Training dataset 1 is used to train the CNN, and training dataset 2 is used to train the stage 2 downstream model; the benefits of using separate datasets are discussed in Sections \ref{incorpweather} and \ref{gen}. The nested structure allows for a robust evaluation of ensemble methods compared to a single test set (see Section \ref{ensembling}). 

\begin{figure}
 \centerline{\includegraphics[width=5.5in]{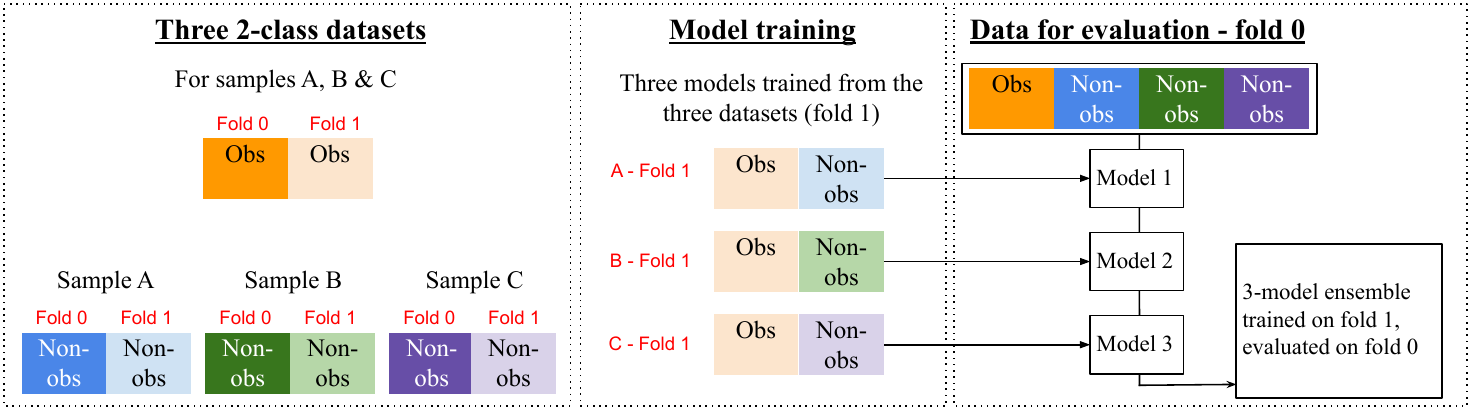}}
 \caption{\textbf{Data splitting for the Obstruction Detection Model}. Sampling, 2-fold CV, and ensembling for the two-class dataset is shown.}\label{fig:ynobssplit}
\end{figure}

For the Obstruction Detection Model, only 164 obstructed images are available. As shown in Fig. \ref{fig:ynobssplit}, three sets of non-obstructed images are sampled from the 5-class dataset to roughly match the non-obstructed count while still representing all classes and sites, yielding 194 non-obstructed images. Due to the small dataset containing 358 samples total, two-fold CV is used to provide training and validation splits only. Although omitting a test set is generally not a best practice, the validation data is used solely for early stopping; no model testing or hyperparameter tuning is performed (see Sections \ref{selection} and \ref{methodsODM}), making two-fold CV a practical choice for this small dataset. The three non-obstructed samples enable a 3-member ensemble: the obstructed images remain the same across datasets, but the varied non-obstructed samples produce three different datasets trained separately for the 3-member ensemble. This allows ensembling without nested CV, which is impractical for such a small dataset.

\section{Modeling methods} \label{methods}

The labeled dataset and its corresponding image and weather data are used to predict road surface conditions across NYS. Classification ML algorithms, including CNNs and RFs, are trained to categorize each observation into one of five road surface classes for the Surface Condition Model. For the Obstruction Detection Model, a CNN is utilized to classify each observation as either obstructed or non-obstructed. Ensembling methods then combine predictions from multiple models trained on different datasets (Sections \ref{datasplitting} and \ref{ensembling}). All models use appropriately normalized and preprocessed input data. Balanced class weighting is used to address the imbalanced dataset (Table \ref{tab:examples_classes}), which is important to encourage the model to learn all classes rather than simply learning to predict majority classes, like wet and dry.

\subsection{Image classification}
Images contain a wide range of colors, textures, edges, and nonlinear, complex patterns. CNNs are commonly used image models because they can automatically extract these complex features in images to learn a classification task. Spatial relationships that exist in the images are learned through filters and convolutional layers \citep{Randy2022}, and nonlinearity in the data is learned by filters and convolutional layers with activation functions.

Images are cropped to remove the top 20\% of the image that often contains text (e.g., timestamps or location, examples in Table \ref{tab:examples_classes}) or is not focused on roadways (e.g., sky or distant scenery) to balance noise removal while preserving essential road surface features within varying camera views. Images are resized to 224x224 and encoded as RGB. Cropping, resizing, and standardization of the data are preprocessing steps that are done on the fly during data loading. The model weights from the epoch achieving highest validation accuracy are saved as the product of training. Table \ref{tab:cnn_config} shows additional CNN modeling details.

\subsection{Surface Condition Model methods}

\subsubsection{CNN models and hyperparameter tuning} \label{cnnbl}
For the Surface Condition Model, a set of 36 baseline CNN models are trained to identify the best performers, of which the best four (according to the criteria listed in Section \ref{selection}) are selected for further hyperparameter tuning. The 36 baseline models are comprised of a combination of six different architectures, three different uses of transfer learning, and the presence or absence of data augmentation.

\begin{table}[ht]
    \centering
    \caption{\textbf{Training configuration summary} for CNN models. Learning rate is indicated as LR.}
    \begin{tabular}{|>{\raggedright\arraybackslash}p{1.7cm}|>{\raggedright\arraybackslash}p{1.7cm}|>{\raggedright\arraybackslash}p{0.8cm}|>{\raggedright\arraybackslash}p{1.0cm}|>{\raggedright\arraybackslash}p{1.1cm}|>{\raggedright\arraybackslash}p{0.9cm}|>{\raggedright\arraybackslash}p{1.1cm}|>{\raggedright\arraybackslash}p{2.2cm}|}
        \hline
        \textbf{Loss} & \textbf{Optimizer} & \textbf{Init. LR} & \textbf{Mom-entum} & \textbf{LR Decay} & \textbf{Batch Size} & \textbf{Epochs} & \textbf{Early Stopping} \\
        \hline
        Categorical Cross-Entropy & Stochastic Gradient Descent & 0.01 & 0.25 & Expon-ential (0.95) & 128 & 30--75 & Val acc. incr.
        $< 0.5\%$ in 10 epochs \\
        \hline
    \end{tabular}
    \label{tab:cnn_config}
\end{table}

Six CNN architectures, which range in complexity, include DenseNet121 from \cite{densenet}, InceptionV3 from \cite{inception}, Mobilenet from \cite{mobilenet}, ResNet50 from \cite{resnet},  VGG16 from \cite{vgg16}, and Xception from \cite{xception}. Data augmentation is used to diversify the training set by applying transformations to the original images, including adjusting the brightness and/or contrast, and horizontally flipping the image. Augmentations are made automatically and randomly to images during model training, with replacement, to make the dataset more robust and varied \citep{tf}.

Transfer learning involves using pretrained models where the backbone of the architecture (which include convolutional layers) is frozen; the weights are already learned from the commonly used ImageNet dataset \citep{imagenet}. Keras provides the architectures and pretrained weights used in this study \citep{keras}. The final layers of the architecture are the dense layers that ultimately result in the final classification, and will be referred to as the ``top” of the model. The top is always trained specifically on the NYSDOT dataset. Two different architecture tops are considered: 1) the generic top (``GEN”) consists of a 2D global average pooling layer followed by three dense layers of size 1024, 512, and six (the output layer), and 2) the architecture-specific top (``AST”) consists of layers that align with the original architecture. A third option excludes transfer learning entirely, training all model weights on the NYSDOT dataset. Thus, three configurations are tested: transfer learning with GEN, transfer learning with AST, and no transfer learning.

For hyperparameter tuning, L2 regularization and dropout are common regularization techniques used to reduce overfitting in CNN models \citep{handsonml}. Regularization is used on the top dense layers, and hyperparameter tuning is completed to identify the best performing L2 weight ($\lambda$) and dropout rate. The hyperparameter space (Table \ref{tab:cnn_hyptunegrid}) includes four non-zero values each for dropout and L2, resulting in 16 sets. Thus, including the baseline runs where there is no regularization (dropout and L2 are 0), there are 17 hyperparameter sets tested in total.

The modeling process begins with running the 36 baseline models, from which the top four are selected using the model selection criteria discussed in Section \ref{selection}. The four models then undergo hyperparameter tuning, and one top performing hyperparameter set is selected for each. Thus, four final CNN models for image classification are considered and carried over into the next step of the modeling process, the downstream classification ML model, which incorporates information from the CNN as well as weather data, as discussed in section \ref{incorpweather}. The final CNN is selected following hyperparameter tuning of the downstream model.

\begin{table}[ht]
    \centering
    \caption{\textbf{CNN model considerations}. Modeling choice sets for baseline runs and hyperparameter tuning (unitless regularization parameters) are shown. \textcolor{red}{The red text indicates the final best CNN model} that is selected and discussed in Section \ref{scm}.}
    \begin{tabular}{|>{\raggedright\arraybackslash}p{1.4in}|>{\raggedright\arraybackslash}p{1.0in}|>{\raggedright\arraybackslash}p{0.7in}||>{\raggedright\arraybackslash}p{0.6in}|>{\raggedright\arraybackslash}p{0.8in}|}
        \hline
        \multicolumn{3}{|c||}{\textbf{Baseline model choices}} & \multicolumn{2}{c|}{\textbf{Hyperparameters}} \\
        \hline
        \textbf{Architecture} & \textbf{Transf. Learn.} & \textbf{Augment.} & \textbf{Dropout} & \textbf{L2 ($\lambda$})\\ 
        \hline
        \{Densenet, Inception, Mobilenet, \textcolor{red}{Resnet}, VGG16, Xception\}
        &
        \{None, \textcolor{red}{AST}, GEN\}
        &
        \{\textcolor{red}{Aug.}, \newline No Aug.\}
        &
        \{0.0, \textcolor{red}{0.2}, 0.4, 0.6, 0.8\}
        &
        \{0.0, 1E-7, 1E-5, 0.001, \textcolor{red}{0.1}\} \\
        \hline
    \end{tabular}
    \label{tab:cnn_hyptunegrid}
\end{table}

\subsubsection{Downstream random forest classification with weather data} \label{incorpweather}
The Surface Condition Model is a two-stage approach that begins with image classification (CNN) and is followed by downstream classification model that is trained on both weather data and the probability output of the CNN model. This stage 2 model is considered ``downstream” because it requires the output from the CNN, and also because it is trained on a separate dataset (training dataset 2) that is \textit{not} used to train the CNN models, as shown in Fig. \ref{fig:nestcv}. The downstream model serves two purposes. First, it incorporates weather data into the modeling process, which is relevant given that the five classes are weather related. Second, by training the model on unseen observations (using training dataset 2), the model is able to adjust for biases in the CNN by learning when the CNN predictions are more (or less) correct and refine its prediction by considering the weather. The improvements from adding weather data are discussed in Section \ref{scm} and shown in Fig. \ref{fig:netgain}. 

First, the CNN model from stage 1 is used to generate predictions (inference) on training dataset 2. Probabilities are calibrated using isotonic regression, a common technique used to create better alignment between predicted probability and accuracy \citep{smooth-isotonic-regression,reliabilitycurves}. Next, the weather data are collected for the time and location corresponding to each image in the hand-labeled dataset (Section \ref{data}). The five probabilities and the six weather variables comprise the 11 inputs for the downstream model, and the hand-labeled classification is the target variable. Multiple classification algorithms and hyperparameters are tested for this step of the modeling process, including dense neural network (DNN), Gaussian Naive Bayes (GNB), logistic regession, Support Vector Machines (SVM), and Random Forest (RF) \citep{chollet_deep_learning, handsonml}. Model details and results are shown in Table \ref{stage2NEW}; the process of selecting models from the set of those tested is discussed in Section \ref{selection}.

\subsubsection{Model selection} \label{selection}

Three parts of the modeling process require model selection: (A) selecting the top four baseline models for hyperparameter tuning, (B) selecting the top hyperparameter-tuned models that are then used in the downstream model, and (C) selecting the final model. All models are evaluated on validation data using the nested-CV setup (Fig. \ref{fig:nestcv}); there are six test datasets with 5-fold CV, totaling 30 different datasets used for training and evaluating each ML model. Results are aggregated across the 30 validation datasets, and model selection is performed by considering multiple metrics that align with the goals of this work while also accounting for intricacies of the imbalanced dataset, as discussed below.

Accuracy is the percentage of correct model predictions (number of correct predictions divided by the number of total observations), and recall can be thought of as a measurement of accuracy \textit{within} a given labeled class (e.g., number of correct snow predictions divided by the number of observations labeled snow). Recall, precision, and F1 score are metrics that are commonly used in ML \citep{handsonml}, but in this study, recall is prioritized because it is the class-specific metric that minimizes the number of misses, which is especially important for this work. As an example, it is crucial for traveler safety that the model does not miss severe snow cases, and recall is the metric that captures this goal. Additionally, since recall measures the performance relative to the number of labeled images in each class, it is an appropriate metric to use with the largely imbalanced dataset. Relying on accuracy alone is insufficient because it is heavily influenced by class imbalance. Performance of \textit{each class} is important for this application, and this is accounted for by calculating the average recall across the five classes (i.e., an average of the five metrics: severe snow recall, snow recall, wet recall, dry recall, poor visibility recall). This average recall metric serves as a more \textit{balanced} performance metric because each class has an equal contribution, rather than accuracy, which is heavily skewed toward the majority classes. Both metrics - accuracy and average recall - are calculated on the validation data and used for model selection by \textit{first} narrowing to the top-performing models based on average recall and then selecting the model with the highest accuracy among them.

\subsection{Obstruction Detection Model methods}\label{methodsODM}

For the Obstruction Detection Model, only a CNN is used because the task is strictly image classification: obstructed or non-obstructed. As discussed in Section \ref{datasplitting}, this model has limitations due to the small dataset with only 164 obstructed examples. With two-fold CV, there is only one validation dataset (no testing dataset), which is needed for model evaluation. Any extensive hyperparameter tuning would \textit{also} require the use of the validation dataset for model selection, which can skew model evaluation results that use the same validation data. Given these limitations, the final CNN architecture and hyperparameters selected for the Surface Condition Model are used to train the Obstruction Detection Model.

\subsection{Ensembling} \label{ensembling}

The data used in this work is variable given the inclusion of many camera sites with varying backgrounds, lighting, and distances from the road. Using one holdout test dataset could lead to performance metrics that do not realistically represent performance if the sites are not perfectly representative of the overall dataset. Thus, for the Surface Detection Model, six distinct test datasets are used, and each one uses 5-fold CV, for which the five models comprise the model ensemble for the outside test dataset, as shown in Fig. \ref{fig:nestcv}. Ensembling combines the outputs from each of the the five models to produce one final prediction. Ensembling is chosen for this application because it can be useful in operations to provide a measure of confidence. Three methods of ensembling are used: (1) average class probabilities, (2) majority vote, and (3) highest-confidence prediction (with calibrated probabilities). Method 1 is used as the prediction when it agrees with either method 2 or 3, otherwise method 2 is used when the prediction agrees with method 3, and lastly, if none agree, the most severe weather condition prediction is used. The same procedure is applied to the obstruction detection model, using a 3-member ensemble (Fig. \ref{fig:ynobssplit}).

\section{Results and discussion} \label{results}

This section includes results for the Surface Condition Model and the Obstruction Detection Model, and it ends with a discussion on generalizability, which is an aspect of this work that is critical for the application to NYSDOT.

\subsection{Surface Condition Model results}\label{scm}

\begin{figure}
 \centerline{\includegraphics[width=5.5in]{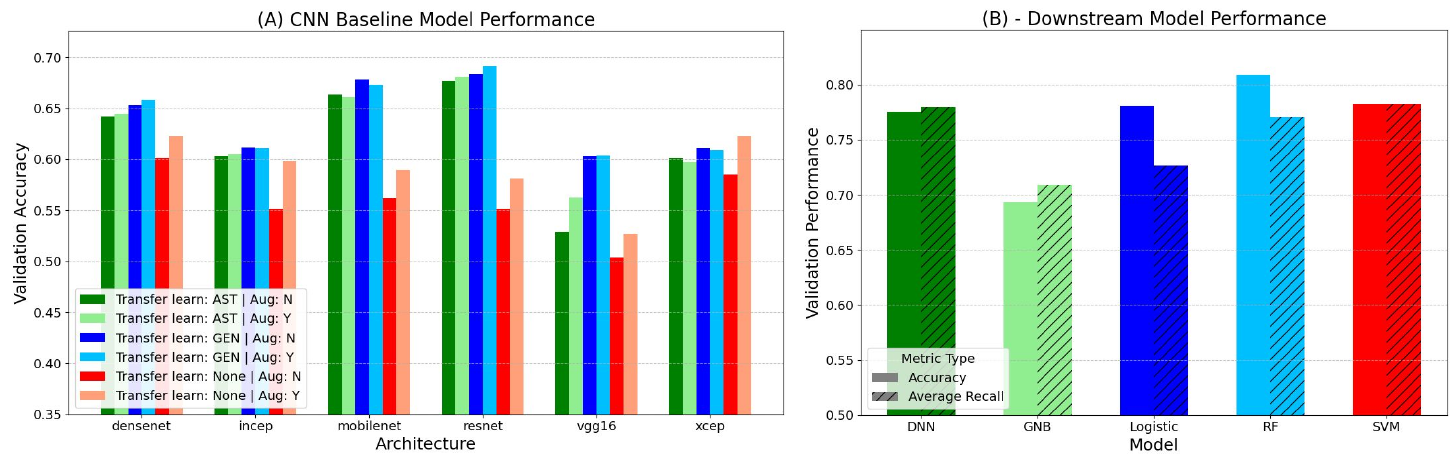}}
 \caption{\textbf{Surface Condition Model: Validation Performance Comparison Across Models}. (A) Validation accuracy of the 36 baseline CNN models are shown; this includes Architecture-Specific Top (greens), Generic Top (blues), and None (reds), and the inclusion of Augmentation (lighter shades), and No Augmentation (darker shades). (B) The top performing hyperparameter set for each of the five algorithms from the stage 2 downstream model are shown, including overall accuracy, as well as the average recall metric. }\label{fig:bl_bar_valacc}
\end{figure}

For the CNN model, there is a 19-percentage point validation accuracy difference between the best and worst baseline models. Densenet, Mobilenet, and Resnet are generally top performing architectures, but the overall performance also depends on the choices of augmentation and transfer learning, as shown in Fig. \ref{fig:bl_bar_valacc}A. Ultimately, the final CNN from model selection (Section \ref{selection}) is a Resnet architecture, using transfer learning with the architecture-specific top (AST), augmentation, dropout rate of 0.2, and L2 regularization rate of 0.1. Hyperparameter tuning and the selection of dropout rate and L2 regularization provide a boost of approximately one percentage point. The selected CNN achieves a validation accuracy of 69.1\%.

\begin{table}[ht]
\centering
\caption{\textbf{Downstream Model Tuning}. Algorithms and their hyperparameters. The validation accuracy (``Acc”) of the top hyperparameter, and the standard deviation (``Std”) and a 95\% confidence interval (``CI”) of accuracy across the 30 unique validation folds are shown for each algorithm. Results are also displayed visually in Figure \ref{fig:bl_bar_valacc}. The \textcolor{red}{red text indicates the final selected model}.} 
\label{stage2NEW}
\renewcommand{\arraystretch}{1.1}
\begin{tabular}{p{0.4in}p{3.5in}p{1.2in}}
\toprule
\textbf{Algo-rithm} & \textbf{Hyperparameters} &
\textbf{Validation results} \\
\midrule
DNN & 
\textbf{Hidden layers}: \{[256,128,64,32,16], [128,64,32,16], [64,32,16], [32,16], [16]\}, \textbf{Dropout}: \{0,0.2,0.4,0.6,0.8\}, \textbf{L2}: \{0,0.001,0.01,0.1\} & \hbox{Acc: 77.5\%} \hbox{Std: 2.4\%} \hbox{CI: (76.7\%, 78.5\%)}\\
\addlinespace
GNB & 
\textbf{var\_smoothing}:\{1e-12, 1e-11, 1e-10, 1e-8, 1e-7, 1e-6\} & \hbox{Acc: 69.4\%} \hbox{Std: 3.8\%} \hbox{CI: (68.1\%, 71.9\%)}\\
\addlinespace
Logistic & \hbox{\textbf{max\_iter}:\{100, 300, 500\},} \textbf{C}:\{0.01, 0.1, 1.0, 10.0, 100.0\}& \hbox{Acc: 78.0\%} \hbox{Std: 2.0\%} \hbox{CI: (77.4\%, 78.8\%)}\\
\addlinespace
SVM & 
\textbf{C}:\hbox{\{1e-3, 1e-2, 0.1, 1, 10, 100\},} \hbox{\textbf{gamma}:\{1e-7, 1e-5, 1e-4, 1e-3, 1e-2, 0.1, 1, 10\}}& \hbox{Acc: 78.2\%} \hbox{Std: 2.1\%} \hbox{CI: (77.5\%, 79.0\%)}\\
\addlinespace
\textcolor{red}{RF} & 
\textbf{estimators}:\{25,100,\textcolor{red}{300}\},\textbf{max\_depth}:\{5,\textcolor{red}{10},20,None\}, 
\textbf{max\_features}:\{\textcolor{red}{3},6,10\},\textbf{min\_samples\_leaf}:\{1,\textcolor{red}{5}\}, \textbf{bootstrap}:\{FALSE,\textcolor{red}{TRUE}\},\textbf{max\_samples}:\{\textcolor{red}{0.5},0.75,1\} & \hbox{Acc: 80.7\%} \hbox{Std: 1.7\%} \hbox{CI: (80.1\%, 81.3\%)}\\
\bottomrule
\end{tabular}

\end{table}

For the stage 2 downstream model that incorporates weather data, five classification algorithms with varying hyperparameters are tested, as discussed in Section \ref{incorpweather} and shown in Table \ref{stage2NEW}. The top performing architecture for each algorithm is shown in Fig. \ref{fig:bl_bar_valacc}B. Results are also displayed in Table \ref{stage2NEW}, including the standard deviation and confidence interval of accuracy across the 30 validation folds. The GNB model has the lowest skill, while the DNN, Logistic, and SVM models all have accuracies around 78\%. The final selected algorithm, following model selection criteria from Section \ref{selection}, is an RF of 300 decision trees with a max depth of 10, a maximum of three features considered in each split, a minimum of five samples per leaf node, and bootstrap sampling of 0.50. This model achieves validation accuracy of 80.7\%.

\begin{figure}
 \centerline{\includegraphics[width=5.5in]{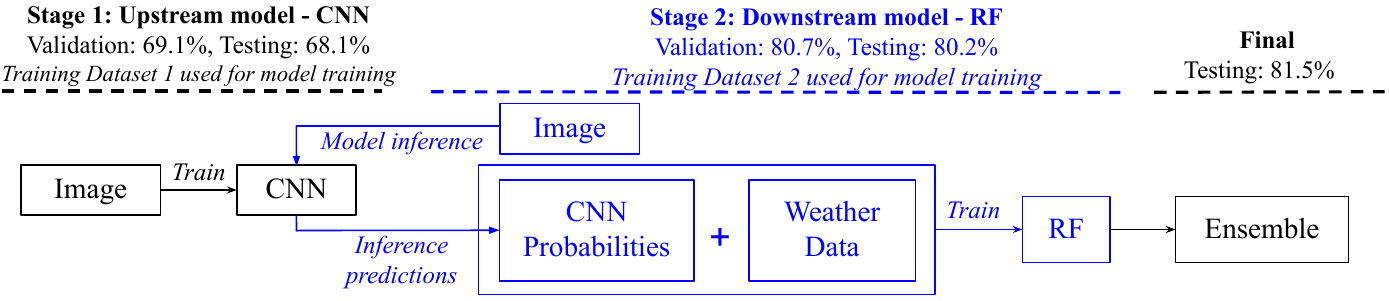}}
 \caption{\textbf{Net Gain in Performance by Step for the Surface Condition Model}. Accuracy on the test dataset is shown for every stage in the modeling process.}\label{fig:netgain}
\end{figure}

The last step in the Surface Condition Model is to get a final prediction for the test dataset by ensembling across the five models (see Fig. \ref{fig:nestcv} and Section \ref{ensembling}). Figure \ref{fig:netgain} shows the test accuracy for every step in the modeling process. Incorporating weather data with the RF provides a significant increase in performance, as shown by the accuracy gain of +12.1 percentage points from stage 1 (68.1\%) to stage 2 (80.2\%). Ensembling provides an additional +1.3\% accuracy increase, resulting in a \textbf{final test accuracy of 81.5\%}. Given the small performance improvement gained from ensembling, deployment of future operational models should consider whether the performance improvement is worth the computational cost associated with ensembling, which creates five times as many models. 

The final model improves predictions compared to the CNN alone, and the most significant improvements are in the snow and dry classes, which increase by 17.3 and 24.3 percentage points, respectively. For the final model (Table \ref{tab:recall}), the classes with the lowest recall are wet, poor visibility, and snow, and this is a reasonable result given that these classes border multiple classes. For example, a simplified progression of a winter weather event is: $\text{dry} \leftrightarrow \text{wet} \leftrightarrow \text{snow} \leftrightarrow \text{severe snow}$. Snow and wet appear in the middle as ``transitional” phases, and therefore have more opportunity for misclassifications, as shown in Table \ref{tab:recall}. Out of all of the model's errors, 95.5\% fell into one of the adjacent classes, which shows that the model makes reasonable errors that align with difficulties that humans also encounter when labeling observations. Similarly, poor visibility is a difficult class given that there are many conditions that can lead to this class, including dense fog, heavy snow, or heavy rain.

\begin{table}[ht]
\centering
\caption{\textbf{Final model performance}. Performance by class for the final model is shown, including recall, precision, F1 score. A one-vs-all approach is used to provide the area under the precision-recall curve for each class (PR-AUC).}
\renewcommand\arraystretch{1.2}
\begin{tabular}{|p{2.35cm}|p{1.5cm}|p{1.5cm}|p{1.2cm}|p{1.2cm}|p{3.4cm}|}
\hline
\textbf{Hand-labeled class} & \textbf{Recall} & \textbf{Precision} & \textbf{F1-score} & \textbf{PR-AUC} & \textbf{Common misclassifications} \\
\hline
Severe snow & 81.5\% & 78.7\% & 80.1\% & .886 & Snow \\
\hline
Snow &  77.1\% & 68.0\% & 72.2\%  & .776 & Wet, severe snow \\
\hline
Wet & 75.9\% & 87.8\% & 81.4\% & .914 &Dry, snow \\
\hline
Dry & 90.1\% & 84.4\% & 87.2\% & .943 & Wet \\
\hline
Poor visibility & 76.8\% & 64.2\% & 70.0\% & .698& Wet, severe snow, dry \\
\hline
\end{tabular}
\label{tab:recall}
\end{table}

As a comparison, the results of \cite{KHAN2022} show a validation accuracy of 95.2\% on a sample of $300$ camera images from different sites, which is higher than the results achieved in this work. This lower accuracy could be due to a variety of factors, including 1) lower camera quality in the NYSDOT images, 2) the inclusion of more road surface classes in the NYSDOT images (e.g. severe snow and poor visibility) and 3) testing on a larger dataset of approximately 3500 held-out images. The importance of data representation is discussed in detail in Section \ref{gen}.

\subsection{Obstruction Detection Model results} \label{resultsODM}

The overall performance for the obstructed class is 88.4\% and for the non-obstructed class it is 97.8\%. Operationally, both the Surface Condition Model and the Obstruction Detection Model could be beneficial to NYSDOT practitioners, thus, all observations would go through both models to provide predictions from each. Model inference of the Obstruction Detection Model on the entire 5-class dataset (21,653) shows that the model correctly predicts non-obstructed on 98.3\% of the labeled 5-class observations, which is even higher than the subset of 194 sampled non-obstructed images that are used in validation (i.e., the 97.8\% recall).

\subsection{Generalizability takeaways}\label{gen}

The choice to have two training datasets - training dataset 1 for the CNN, and training dataset 2 for the RF - inherently aligns with the goals of generalizability because it incorporates training from sites that are unseen from the CNN. The value in doing this is proven further by running an experiment where the RF and CNN are trained on the \textit{same} dataset, using all four training folds from Fig. \ref{fig:nestcv}. The validation skill decreases 2.2\% (from 80.7\% to 78.5\%), supporting the usefulness of training on a downstream model. Additionally, feature importance analysis is performed to explain the contribution of each input variable in terms of model skill. When the CNN and RF are trained using the same dataset, the contribution of weather data is only 21\%, whereas the contribution is higher, 34\%, when models are trained on separate datasets. This result supports the decision to train on separate datasets because it encourages more balanced contribution from weather and image data.

Another experiment is run to confirm the usefulness of incorporating weather data for generalizability. For this experiment, an alternative shuffle-split method is employed where each site is represented in every fold. This method has higher validation accuracy, which is expected because the model is trained on every site that exists in validation, and is not applicable for the operational end goals of this work. With the shuffle split method, validation accuracy for stage 1 (CNN) is 80.8\% and stage 2 (RF) is 86.8\%, which is an accuracy increase of 6 percentage points. However, the site-specific split method has a validation accuracy increase of +11.6 percentage points (69.1\% to 80.7\%, as shown in Fig. \ref{fig:netgain}), which is nearly double that of the shuffle split method. This comparison indicates that the incorporation of weather data downstream plays a critical role in improving model generalizability.

The importance of site representation is demonstrated with an additional experiment where half of the camera sites are removed from the labeled dataset. The same sites are removed from both the shuffle split and site-specific split, and the halved datasets contains the same proportion of both high-quality and low-quality sites compared to the full dataset. The purpose of conducting this experiment is to answer the question of whether more data, specifically more data from \textit{more sites}, is helpful for generalizability. For the site-specific split method, results show that halving the dataset reduces model accuracy by approximately -3.2 percentage points (from 80.7\% to 77.5\%), while the shuffle split only has a reduction of approximately -0.1 percentage point (from 86.8\% to 86.7\%). This indicates that more data from new camera sites is particularly helpful for generalizability because the site-specific method sees a greater benefit from the inclusion of more sites. Furthermore, low quality sites (examples shown in Table \ref{tab:examples_classes}) especially benefit from the addition of more sites with an accuracy increase of +6.4\%, whereas high-quality sites see an increase of only +1.8\%.

Overall, low quality sites perform worse than high-quality sites, with low-quality sites having validation accuracy of 74.1\% and high quality with 85.5\% (with overall validation accuracy of 80.7\%). This means that a task may be made more difficult by the inclusion of low-quality sites. This performance contrast is intuitive, but it raises interesting considerations for dataset curation and model development. Site representation is critical given the goal for statewide deployment, but it is something that can easily be neglected due to the difficulty of labeling poor-quality sites, as well as lower performance metrics associated with lower quality sites. However, it is important to recognize that, in this context, lower accuracy does not necessarily mean it is a ``worse” model because the end goal of this work is to develop a model that can be applied on cameras across NYS, and those cameras \textit{do} range in quality. Deploying a model heedlessly without considering representation could result in an inapplicable operational model. End-users, data curators, and model developers must work together to ensure the application is being met.

\section{Conclusion and future work}
A large network of camera images and weather forecast data are used to predict weather-related road surface conditions with 81.5\% accuracy on unseen test sites. An Obstruction Detection Model is also developed to predict obstructed image instances, correctly classifying 88.4\% of obstructed cases. For operational application, the full pipeline - data preparation and model inference - runs in approximately 3.5 minutes for all camera sites statewide on a V100 GPU with 8 CPUs and 64 GB of memory, demonstrating feasibility for real-time updates every five minutes.

This study emphasizes generalizability to new and unseen sites, which is critical for operational deployment by the NYSDOT and other agencies. The inclusion of weather data proves to be particularly beneficial for increasing model generalizability through an increase of +12.1\% accuracy. Including additional data from new and diverse sites further enhances generalizability, especially for low-quality cameras. The inclusion of low-quality sites and difficult classes like poor visibility make the task more difficult but are important for realistic operational deployment. The focus on generalizability, along with the robust data curation methods developed for this work \citep{Sutterzenodo, Sutterdatasheet}, provides a framework that can be extended to other agencies and locations. 

Future work should include operationalization of the tool for NYSDOT through creation of a dashboard or API, including feedback from end-users to assess and improve the usefulness of the tool. Collaboration with the NYSDOT has been and will continue to be critical for the adoption and usability of this work. Expansion to other regions and states should be explored, including the addition of more labeled data. From a modeling perspective, future work should consider post-processing techniques, inclusion of additional weather variables, and testing of other modeling techniques such as recurrent neural networks, co-training or discriminative self-supervised learning \citep{rothenberger2024meta}. Overall, the models developed in this work can support frequent (every five minutes) and wide-scale monitoring of road surface conditions at the camera level, offering information valuable for resource allocation and driver safety during hazardous winter weather events.

\printcredits

\section*{Data availability statement}
The dataset is private, but comprehensive information is provided in \cite{Sutterdatasheet}. The code is available at https://github.com/cgsutter/DRIVE-clean.

\section*{Declaration of competing interest}

The authors have no known competing financial interests or personal relationships that could have appeared to influence the work reported in this paper.

\section*{Acknowledgments}
This material is based upon work supported by the U.S. National Science Foundation under Grant No. RISE-2019758. This work is part of the NSF AI Institute for Research on Trustworthy AI in Weather, Climate, and Coastal Oceanography (NSF AI2ES). \cite{xcite} provided computational resources for model development. This research is also made possible by the New York State (NYS) Mesonet. Original funding for the NYS Mesonet (NYSM) buildup was provided by Federal Emergency Management Agency grant FEMA-4085-DR-NY. The continued operation and maintenance of the NYSM is supported by National Mesonet Program, University at Albany, Federal and private grants, and others. 

\bibliographystyle{cas-model2-names}

\bibliography{references}





\end{document}